# Learning Pose Estimation for High-Precision Robotic Assembly Using Simulated Depth Images

Y. Litvak, A. Biess*, A. Bar-Hillel*

*Abstract*— Most of industrial robotic assembly tasks today require fixed initial conditions for successful assembly. These constraints induce high production costs and low adaptability to new tasks. In this work we aim towards flexible and adaptable robotic assembly by using 3D CAD models for all parts to be assembled. We focus on a generic assembly task - the *Siemens Innovation Challenge* - in which a robot needs to assemble a gear-like mechanism with high precision into an operating system. To obtain the millimeter-accuracy required for this task and industrial settings alike, we use a depth camera mounted near the robot's end-effector. We present a high-accuracy two-stage pose estimation procedure based on deep convolutional neural networks, which includes detection, pose estimation, refinement, and handling of near- and full symmetries of parts. The networks are trained on simulated depth images with means to ensure successful transfer to the real robot. We obtain an average pose estimation error of 2.16 millimeters and 0.64 degree leading to 91% success rate for robotic assembly of randomly distributed parts. To the best of our knowledge, this is the first time that the *Siemens Innovation Challenge* is fully addressed, with all the parts assembled with high success rates.

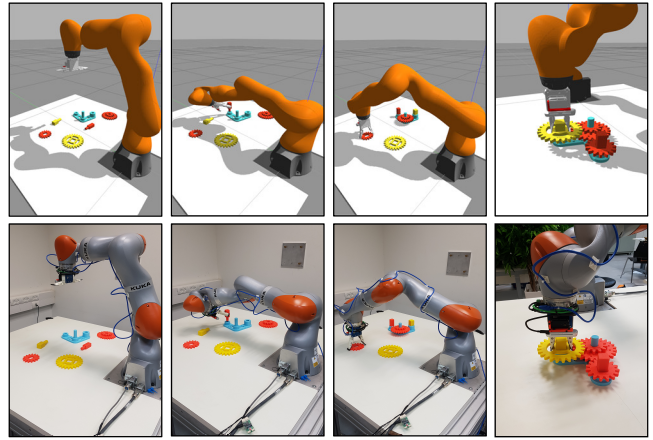

**Figure 1**: The KUKA LBR IIWA robot performs the *Siemens Innovation Challenge* successfully in simulation (top row) and in reality (bottom row).

## I. INTRODUCTION

Robots in manufacturing and assembly tasks are often lacking perception, making it necessary to place the parts to be manipulated in pre-defined positions using fixtures or part feeders. The installation of such structured environments is a time-consuming process, which induces high costs. To enable the next advance in robotic manufacturing, where a large variety of products is likely to be manufactured in small production volumes (Industry 4.0), robots need to be endowed with adaptive and flexible skills. Flexibility here means the ability to work with *arbitrary* initial part conditions. Our goal is to obtain flexible assembly based on recent advance in machine learning, computer vision and depth sensing.

Many studies involving manipulation of parts are proposing methods for grasping from arbitrary initial condition [1-10]. Some consider kitting and basic assembly tasks [3,11,12], like stacking of Lego bricks [12] or peg-in-the-hole problems [11]. Most of the grasping studies consider grasping of *general* objects, which is considerably more difficult than grasping of a known specific part. In addition, most of these studies consider RGB images obtained with a camera in a fixed position covering the entire scene, and some of them train mappings directly from images to actions. This creates highly diverse image distributions, leading to difficult learning problems. The cost of learning in such large input spaces is often reduced accuracy (the ability to bring a part exactly to a target pose) and reduced robustness (the probability to accomplish the task successfully). When all the parts of an assembly task are rigid and known a-priory as CAD models, a different approach can be used, which we pursue here. First, the task can be completely defined in terms of start and end target poses of the parts. The target poses are naturally defined as relative part poses (pose of one part in the reference frame of the other), but can trivially be translated to the robot reference frame once the poses of all parts are estimated. In addition, the end effector pose required for a successful grasp of each part can be stated a-priory (in the part's reference frame), as part of the task definition. This implies that an assembly task can be primarily reduced to pose estimation, provided that motion control is sufficiently accurate.

An appealing approach for improving pose estimation is by making the learning problem easier: use a small input space, and a large training sample. A large sample is difficult to obtain on a real robot [2], but can be easily generated in a simulated environment [4-7,10,11,13]. When a pose estimation algorithm is trained in a simulator, adapting it for execution on the real robot may be challenging, especially for RGB images and wide scene view. In recent years techniques for bridging the simulation-reality gap in RGB images were proposed like domain randomization [4, 5] or images synthesis with GANs [4]. However, bridging the simulation–reality gap is still difficult with significant accuracy reduction involved. We hypothesize that bridging the gap is simpler for synthetic depth images, which are more similar to their real counterparts than synthetic RGB images [7].

These considerations led us to the following working assumptions in our approach: First, it is assumed that exact 3D

* These authors contributed equally, same affiliation for all authors.
Research supported by the ABC Robotics Initiative, Helmsley Charity Foundation. Y.L. is with the Department of Industrial Engineering and Management, Ben-Gurion University of the Negev, Be'er Sheva 8410501, Israel.(corresponding author, phone: +972-52-523-4050 email: litvaky@post.bgu.ac.il) A.B. (email: abiess@bgu.ac.il), A.B-H. (email: barhille@bgu.ac.il)

CAD models of all rigid parts of interests are available. Second, we rely exclusively on depth sensing. Third, the depth sensor is mounted next to the robot's end-effector in order to place the end-effector in a canonical pose as near as possible to the part of interest. Placing the camera as close as possible to the object is a key for pose estimation accuracy, since the world pose estimation error is determined by a multiplication of the error obtained in the image plane and the distance between object and camera focal point [14]. The importance of small sensor-object distance is enhanced by the fact that current of-the-shelf depth cameras are of relatively low resolution compared to available RGB cameras. Since CAD models for all parts are available, all training is done with simulated data only. A policy for a new task may then be trained within several hours of computation. Hence, the approach – if successful – has the potential to be adaptive by enabling fast migration to new tasks.

In this study we test the suggested approach on a single assembly task and try to solve the *Siemens Innovation Challenge* (https://www.siemens.com/us/en/home/company/fairs-events/robot-learning.html). In the challenge a robot needs to assemble a gear-like system composed of a base plate, two different shafts and three different gears into an operating mechanism (Fig. 1). The task can be broken into 5 sub-tasks, each requiring grasping and assembly of a single part. The sub-tasks are highly challenging since they require millimeter accuracy in positioning of the parts. The initial poses of the base plate and the parts to be assembled are not constrained and arbitrarily placed without overlap in the workspace of the robot. In addition, some parts have rotational symmetries, or near-rotational symmetries, of different orders. The *base plate* has no symmetry (or symmetry of order *n=1*). *Gear 2* has rotational symmetry of order *n=4* (i.e., rotations of 360/4 degrees lead to the same shape) and the *compound gear* has rotational symmetry of order *n=12*. *Gear 1, shaft 1* and *shaft 2* have near-symmetries, meaning rotations of the part resemble each other, but are not identical (see Fig. 2 and 4).

For a rigid part, the pose in 3D is fully defined by six parameters (three translational and three rotational degrees of freedom). However, for many robotic applications, and specifically for the task we consider, it is sufficient to describe the pose of a part in the plane, which requires only three parameters (two translational and one rotational degree of freedom [8,15]). The main component in our approach is the pose estimation pipeline, composed of two stages implemented as Convolutional Neural Networks (CNN). In stage one, several images covering the scene are taken from a pre-defined height above the workspace surface. These are processed by a RetinaNet [16] based detector and a pose estimator, providing detection and initial pose estimation for each part. In stage two, the robot's end-effector is moved to take a close-up view image of each part separately. The image is taken in a canonical pose based on the current pose estimation, so the part is expected to appear in the image at a certain known position, size and rotation. This image is then used for further pose refinement and resolving of near-isometry ambiguities. The high resolution and limited appearance space due to the enforced canonical pose enable efficient learning of accurate pose estimators. While stage one is executed once to provide initial pose estimations for all parts, stage two is executed separately for each part at the beginning of its assembly.

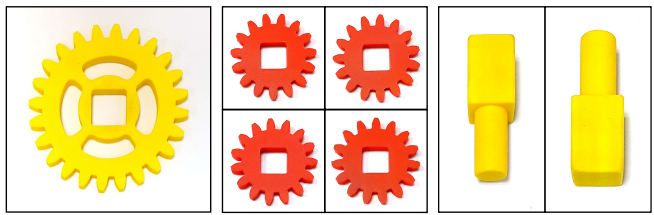

**Figure 2**: Isometry and near isometry of real assembly parts: (Left): *Gear 2* part with rotational symmetry every 90 degrees. (Middle): near-symmetry views of the *gear 1* with 4 near-symmetries rotated to canonical pose. Note that while these four rotations of *gear 1* are very similar, successful classification among them is required for exact pose estimation. When another rotation is used instead of the correct one, the grasp efficiency will be reduced due to lack of contact points and the gear's teeth may not fit well into the teeth of its neighboring gears. (Right): two near-symmetries of *shaft 2* rotated to canonical pose, which are much easier to discriminate.

Our environment consisted of a KUKA LBR IIWA 14 R820 anthropomorphic robotic manipulator equipped with a SAKE ROBOTICS EZGripper and an Intel RealSense D415 depth camera mounted on the gripper (Fig. 1). For each part, except of the *base plate*, the task definition includes two relative positions: a final required pose relative to the *base plate*, and a grasp pose (of the end-effector) defined in the part's reference frame. A script encoding the assembly sequence was written, including a grasping and assembly motion for each part with trajectories computed using standard motion planning. In addition, a force-based feedback routine is employed upon each assembly part if part misalignments are detected by using a threshold on the force sensor reading. Assembly of the part is then re-attempted in multiple offsets around the expected position.

Our experiments included 58 assembly attempts on the real robot. In these experiments part detection rate was 100%, near isometries were resolved in 100%, the mean translation error of a part was 2.16mm, and the mean rotational deviation was 0.64 degrees. This accuracy enabled successful grasp in 97.9% of the cases, and successful assembly in 91% of the cases. While direct comparison is not possible, the obtained accuracy and the scope of successful assembly is higher than in previous related work [5,11]. Most of the failures are related to the grasp instability due to a sub-optimal gripper: the fingers cannot reach a parallel grip in certain cases and the force applied by the gripper is not strong enough. We hence believe the suggested approach is a significant step towards flexible and adaptive CAD-based robotic assembly.

## II. RELATED WORK

We briefly discuss topics which are mostly related to our work: flexible assembly and part manipulation, pose estimation, and exporting policies from simulation to reality.

*Flexible Assembly:* Recent work has often focused on grasping from arbitrary initial conditions ([1,2,4-10] are a small subset). The interest is often in grasping of general unknown objects (i.e. objects not seen in training), for which very high success rates can be obtained when large simulated [10] or real datasets [2] are used. However, high-accuracy manipulation problem addressed here is only partially similar to grasping and is orthogonal to it in several senses. On the one hand, the networks we train do not solve general grasping, but only grasping of specific known objects. On the other hand, our assembly manipulation task requires to go beyond traditional metrics of grasp success rates. We have to grasp objects with very specific relative poses (of the gripper w.r.t the object), which enable proper assembly with millimeter accuracy which is usually not required for successful grasping.

Beyond grasping, several works have considered manipulation and assembly tasks [3,11,12]. These tasks include grasping and bringing parts to tight spatial relations with others. Several manipulation tasks have been learned on a real robot, like placing a coat hanger on a rack or screwing a cap on a bottle [3]. The learning task is split between two agents: a mixture of linear Gaussian controllers, each trained for specific known initial conditions and a deep network generalizing across initial conditions. In [11] the tasks are most similar to ours, as they use a subset of the *Siemens Innovation Challenge* tasks (two sub tasks), as well as a U-shape fitting task. In this work the pose estimation problem is addressed using fiducials (attached markers on the parts designed for pose estimation). Training is done using a combination of motion planning and Guided Policy Search (GPS), where the former is used to guide training of the latter. High success rates are reported for the real robot, but only from two known and fixed initial part positions. In [12] a difficult assembly task of Lego brick stitching is handled, but only in a simulation environment. The authors assume known states, including brick positions, and use a DDPG algorithm with several improvements to learn assembly policies.

The studies in [3] and [11] address a difficult learning problem of complex observation-to-action maps. Our approach is inherently different: We avoid the need for learning observation-to-action maps by enabling highly accurate state (pose) estimation, from which simple motion planning suffices for task accomplishment. We obtain this accuracy by using two main ideas: first, we narrow the input distribution to face a relatively easy learning problem by moving the sensor to the vicinity of the part with a relatively fixed spatial relation and use of near-range depth imaging. Second, we solve this problem in the simulator where a very large sample can be used to learn a complex pipeline of multiple refining CNNs with high accuracy.

*Pose estimation*: Pose estimation of an object starts with object detection, for which significant progress has been made in recent years [16,17,18]. In the recent vision literature, pose estimation is often performed for non-rigid objects like the whole human body [19] or the human hand [20]. In these cases, the pose is specified using the location of multiple interest points. In contrast, the pose of a rigid object in 3D is fully defined by six parameters, or three if only pose in the plane is considered. In the Cornell grasping data set [15] the grasping task is defined by specifying the 2D-pose of the gripper (not the object) required for a successful grasp. Several network architectures were suggested for the regression of successful grasp poses, including a two-stage process of candidates finding followed by candidate ranking in [15] and regression in multiple spatial cells in [8]. Other CNN architectures for pose estimation in the robotic context include networks with spatial soft-max for feature point finding [2] or convolutional pose machines [19], where multiple layers regress the same heat map for refinement.

A possible approach to reduce the complexity of pose estimation is achieved by using depth information [7,13]. Specifically [7] uses a depth camera mounted near the gripper as is done in our work. In our study we combine the strengths of [7] with a process of canonization and refinement, similar to [20]. When depth information is present, pose estimation can be addressed without learning, by registration of a 3D model and the observed point cloud using methods like Iterative Closest Point (ICP) [21]. Such methods can be rather accurate, but they provide local optimum of the registration and require good initial registration hypothesis. In a task like the *Siemens Innovation Challenge*, where some parts have strong self-similarity due to near-isometry, it may be very difficult to find a good initial hypothesis, which will lead to the right registration and not result in a wrong local optimum.

*Simulation to reality transfer:* When a predictor is defined over RGB images, transferring between simulated and real images is challenging due to significant differences between simulated and real images in terms of illumination, texture and background distribution. In domain randomization [4,5] scenes are generated under a large variety of visual conditions (e.g. illumination conditions and background textures), object types and robot dynamics characteristics. A predictor trained on such a rich distribution is usually more amenable for transfer to the real world. Other domain adaptation techniques are domain–adversarial neural networks (DANN [22]) and a transfer GAN creating seemingly-real images from simulated images [5,23]. Following [13,24], we hypothesize that the gap between simulation and reality is smaller and easier to bridge when only depth images are used. Specifically in [13], a 6D grasp pose detector has been trained in a simulator and transferred successfully to the real robot with 93% success rate without any special mechanisms used for transfer. In [24] it was shown that by enhancing simulated data to mimic the signal processing pipeline of a depth camera transfer can be further facilitated. We follow some of these ideas in this work.

III. METHOD

*A. Pose Estimation – General Overview*

Our pose estimation pipeline includes two stages: (1) detection and coarse pose estimation from a distal point of view and (2) high-precision pose estimation and classification of near-symmetries from close-up views.

*Stage 1 - Detection and coarse pose estimation:* Several depth images covering the entire surface are taken from a distance of 0.53m above the table. These images are analyzed for parts and poses using a fully convolutional neural network (CNN) with architecture similar to the RetinaNet [16]. For each detected part the network outputs the part's class (one of six known classes), a bounding box around the part, and the orientation angle $\theta$ of the part around the $Z$ axis of the world frame. The real position of the part's center on the workspace surface is then calculated by de-projecting the center pixel of the estimated bounding box. To simplify learning, the range of the orientation angle $\theta$ for each part is defined by the symmetry of the part spanning an interval from 0° to *360/n*, where *n* is the order of the rotational symmetry. For symmetric parts, *compound gear* and *gear 2*, the maximum angles are 30°

(*n=12*) and 90° (*n=4*) respectively. For near-symmetric parts, *gear 1*, *shaft 1* and *shaft 2*, the maximum angles are 90°, 180° and 180° respectively.

*Stage 2 - High-precision pose estimation from close-up views*: To obtain high-precision pose estimation results, we analyzed images from close-up views. For this purpose an additional depth image of each part is taken from a distance of 0.31m (the closest distance for which the largest part can be fully seen) with the camera rotated into canonical pose according to the estimation obtained from stage 1. The output of stage 2 is a–bounding box estimate, which enables good prediction of the part center in the world frame (as done in stage 1), a fine-tuned correction for the orientation angle $\theta$ and a sub-class classification for the near-symmetric parts: *shaft 1, shaft 2* and *gear 1*. For shafts, there are two possible sub-classes: the part is close to 0° (class 0) or close to 180° (class 1). Respectively, for *gear 1* four sub-classes are possible, and the distinction among them is more difficult (see Fig. 2.center). The final part rotation is determined based on the near isometry classification (for determining the baseline angle) and adding the regressed orientation correction.

We have also experimented with a three stage system, in which an additional stage was added between stage 1 and 2. In this stage each part detected at stage 1 was cropped from the original image and rotated into a canonical pose by the estimated (negative) orientation angle $\theta$. A network trained on such cropped images was used to refine the pose estimation and resolve near-isometries before taking close-up images (to improve the canonical pose in which they are taken). However, our experiments have shown that the three-stage method did not obtain higher accuracy than the two-stage method.

*B. Pose Estimation – Networks Architectures*

*Stage 1*: Our network is similar to the RetinaNet architecture, with architectural modification made for our specific needs (Fig. 3). The network's backbone in stage 1 is ResNet-50 [25], a residual network pre-trained on ImageNet dataset, followed by a Feature Pyramid Network (FPN) [26], which is effective in detection of multi-scaled parts. The FPN extracts features from three internal layers of ResNet-50 to construct a pyramid of five layers, *P3-P7*. These layers contain similar representations, summarizing features from high and low layers of the ResNet backbone, but in five different spatial resolutions. Even though the images in our environment are taken from fixed heights, RetinaNet's backbone is used without modification. Anchors for object detection were used as in the original network with five octaves and nine anchors at each point, representing different scales and aspect ratios. During training anchors are labeled as positive, negative or 'ignore' based on the intersection-over-union (IOU) between the anchor and ground truth bounding boxes, with original thresholds set to 0.5 and 0.4 for positive and negative overlap, respectively. Since we aim towards more accurate detection, we set the thresholds to 0.7 and 0.6, respectively.

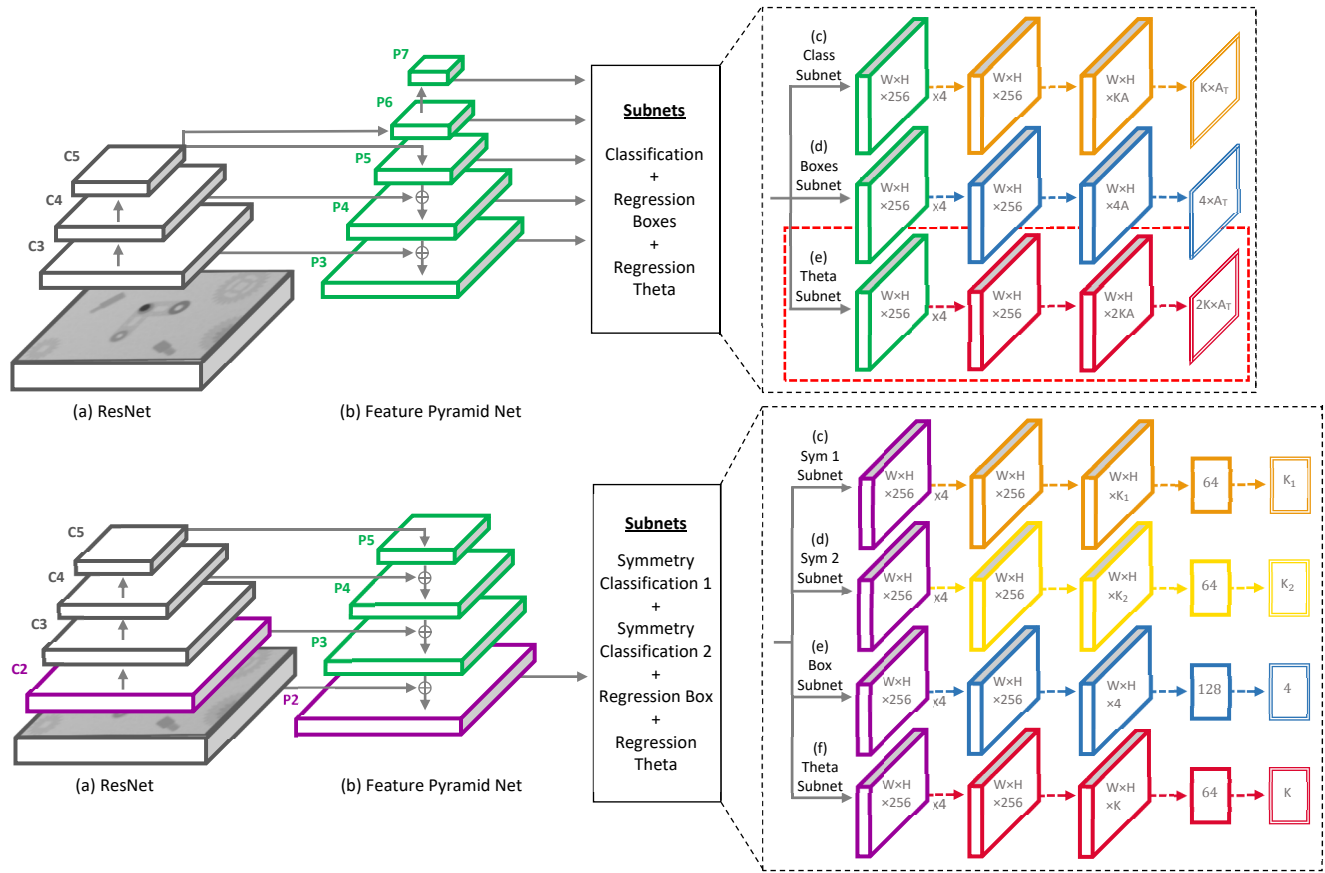

**Figure 3**: Networks' architecture. (Top graph): Stage 1 network, based on RetinaNet architecture, starting with a backbone of FPN on top of a feedforward ResNet (a-b), followed by three subnetworks. First two subnetworks are classification (c) and regression of bounding boxes (d) and the third and new subnetwork, marked with a red square, is regression of rotation angle $\theta$ (e). In this drawing $A = 9$ is the number of anchors per location, $A_T$ is the total number of anchors in the image and $K = 6$ is the number of part classes. (Bottom graph): Stage 2 network. We use four internal layers from ResNet (a) to produce a high-resolution layer from the FPN (b), followed by 4 subnetworks (c-f). $K_1$ is the number of near-symmetries classes learned in Symmetry 1 subnetwork (c), $K_2$ is the same for (d), and $K$ is the total number of classes ($K_1 + K_2$ + all other parts which don't have near-symmetries). The last layer in each subnetwork is the output layer.

RetinaNet includes two sub-networks: one for classification and one for regression of the bounding boxes. The losses for the first two sub-networks were used without change: Focal loss [16] for part classification, and Huber loss (smooth $L_1$) for bounding box regression [17]. A third sub-network for regression of the orientation angle $\theta$ was added in stage 1. The third subnet of orientation regression outputs for each [anchor, part] the orientation $\theta$ parametrized as $(\sin(n\theta), \cos(n\theta))$, where $n$ is the rotational symmetry order of the part (following [8]). Since for most part-anchor pairs no part is detected, no loss is applied in such locations during training – the loss only applies in true detection anchors and only for the relevant part. We use the Huber loss over the two relevant orientation neurons for each class.

*Stage 2*: This stage does not perform detection, as input images with a single part are provided to the network. We use the FPN technique to create a high-resolution map (twice the size of the largest map used in stage 1) termed $P2$, based on the $C2$ ResNet layer. This map contains rich features from higher ResNet layers, yet in a high resolution enabling fine grained distinctions. This layer is the input of four sub-networks. The first two classify parts with near-symmetries. Subnet 1 predicts the four possible orientations of *gear 1* and subnet 2 the two possible orientations of *shaft 1* and *shaft 2* leading to a total of four outputs. The loss for these two subnets is the standard categorical cross entropy. Like before, the subclass loss is only used in training for parts in the relevant class (*gear 1* for subnet 1, shafts for subnet 2). The third subnetwork regresses a refinement of part's bounding box. The subnet has only 4 outputs, since there are no multiple anchors. The loss function is the Huber loss. The forth subnetwork regresses the orientation angle $\theta$, similar to the one in stage 1. Since the part is almost in canonical position and the predicted angular range is within [-10, 10] degrees, no cyclic angle needs to be considered here, so the output neurons directly regressing $\theta$ (one neuron for each class).

In all subnetworks of stage 2, a fully connected layer with 64 or 128 neurons is added after the last 2D-convolutional layer and before the output layer of the subnet, to improve the accuracy of the network. This has a slight cost in computation time, which was not optimized in this work. In the classification subnets a *ReLU* activation function was used after the fully connected layer, whereas in the regression subnets no activation function was used.

### C. Creating Training Dataset

Our simulated environment was built in Gazebo [27] – a ROS simulation software – and consisted of a KUKA LBR IIWA simulated robot [28], to which we added the gripper, camera, table and parts. Simulated depth images were created with the Gazebo depth camera plugin.

*Stage 1*: Depth images of randomly distributed parts were taken from a distance of 0.53m above the table. Due to the limited field of view of the camera only some of the parts are visible in each image, but randomization parameters were adjusted such that the total number of appearances of each part is similar. Each image contains 1 to 5 parts, with an average of 3.2 parts per image. 200,000 images were created for training with proper ground truth. The bounding box ground truth was determined by the projected center of the part in the image and the part's radius. For each part, the bounding box size is fixed and does not depend on the part orientation. To estimate the ground truth rotation angle in stage 1, all near-symmetric parts were treated as symmetric. For example, if *gear 1* is placed in an orientation of 215 degrees, the ground truth angle in stage 1 would be mod(215,90) = 35. This value is then multiplied by the symmetry order to create the cosine and sine ground truth.

*Stage 2*: Depth images were taken from a distance of 0.31m above the table (in the simulation environment). For each image, only one part was placed on the table. Each sub-class of the parts with near-symmetries was handled separately. 50,000 images were created for training of each sub-class, leading to a total of 550,000 images. In preparation, images of the parts were generated in perturbed image positions $(\hat{x}, \hat{y}, \hat{\theta}) = (x, y, \theta) + (\delta x, \delta y, \delta\theta)$. Here $(x, y, \theta)$ is the ground truth position, and $(\delta x, \delta y, \delta\theta)$ is a random noise vector with its components drawn independently from uniform distributions with small $\sigma$ parameters. The ground truth for this stage includes sub-classes of near-symmetries, $\delta\theta$ for the rotation refinement subnetwork, and the perturbed bounding box of the part for the bounding box refinement subnetwork. Hence the network is trained to identify the part misalignment (and sub class when relevant) in a population of images containing roughly centered parts.

### D. Bridging the simulation-reality gap

To make the simulated depth images more similar to real depth images, the synthetic images were processed with a few steps attempting to mimic the noise and signal processing pipeline of a depth camera (Fig. 4). First, an additive Gaussian independent pixel noise of standard deviation $\sigma$ was applied to all image pixels. For each image $\sigma$ was chosen uniformly in [0.5,3] cm. Although the reported noise per pixel of the real D415 depth camera is only 1cm, we use this procedure for the system to be resilient to various noise conditions, which may result from variance in illumination, camera-object distance, etc. Second, a Gaussian filter with randomized standard deviation $\sigma$ was applied to the image. A similar function is often used in the real depth camera processing pipeline, but with an unknown $\sigma$ parameter. We hence randomize $\sigma$ uniformly in [2,5] pixels. Following these operations, the minimum and maximum depth values were limited to a certain range and then values were stretched to the range of [0,255]. The 2D images were finally expanded to three identical maps creating a 3D RGB greyscale image required as input by the RetinaNet network. In addition to this operation of bringing synthetic images closer to real, we take some actions to bring real images closer to synthetic. See section IV.A for details.

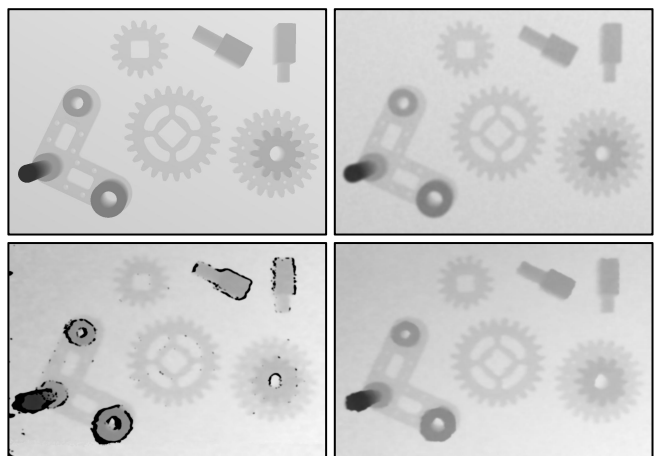

**Figure 4**: Depth images, showing all 6 parts, taken from distance of 0.53m in simulation (top row) and in real (bottom row), before image processing (left column) and after (right column). Parts' classes by order of appearance (from left part and clockwise): *base plate, gear 1, shaft 2, shaft 1, compound gear* and *gear 2*.

*E. Assembly*

For each part assembly, a sequence of pick-and-place operations is computed based on the estimated poses. The pick operation requires the grasping pose of the robot's end-effector with respect to the part, whereas the place operation is determined by the part's pose with respect to the *base plate*. The required pose for each operation is computed based on the estimated part and *base plate* positions. Calculation of target poses include simple homogeneous matrix multiplications. For example, the pose of the end-effector in world coordinates is obtained by the pose of the part in world coordinates multiplied by the pose of the end-effector with respect to the part. Based on the current and target pose of the end effector, motion trajectories are computed using Moveit! [29] – a ROS software motion planning package. The planner generates a suitable trajectory in the 7D joint-angle space of the robot considering all of the robot's constraints. Thus, our assembly process consists of estimating the pose of the parts from real world images followed by applying a sequence of part pick-and-place motions calculated with Moveit!.

To handle estimation errors, a force feedback routine is added and applied to the assembly of each part. All the parts are assembled using a descend motion bringing the part to its proper position. If during the last descend of the place operation the robot's force sensor measures a significant force in the opposite direction (i.e. up in the world Z direction), an unexpected contact with the *base plate* or another part is indicated. This, in turn, shows that the part was not placed in the correct position, and the robot stops the descend motion. This event triggers a search algorithm for a new descend position starting from positions which are closer to the initial estimate and going outwards in circles until a position with no negative force is found (Fig. 5). To find the array of possible offset locations, we think of the problem as finding a minimal set of positions, which guarantee successful insertion of a disk of radius $r$ (for example: a shaft) into a hole of radius $R$ ($r<R$) located in an unknown position within a squared area. The search locations of an optimal set for this problem are arranged in equal triangles with side lengths $d = 2(R-r)$ and can be found using a known algorithm [30].

## IV. EXPERIMENTS

*A. Pre-processing*

Real depth images have spatial noise effects (in addition to independent pixelwise noise) and zero-value pixels, mostly near large depth gradients where the camera has difficulties to obtain good depth estimates. To remove some of the noise, we always capture ten copies from the camera for each image and calculate the average of all none-zero pixels. All remaining zero-pixels are replaced by the average of their nearest neighbors. Finally, we use the same minimum and maximum depth limits defined during the creation of the simulated images and transform the image to a 3D RGB greyscale image (Fig. 4).

The parts of the *Siemens Innovation Challenge* were printed using a 3D printer. For assessing the quality of our pose estimation procedure, we had to create part configurations with known ground truth of the parts' poses on the table. For this purpose, the parts were randomly placed in simulation and their poses were saved to a file. At the beginning of a real experiment, each part was manually placed in the corner of the table – a fixed known position – and then transferred by the robot to its previously saved position on the table.

During preparations, we discovered that our gripper was not centered and aligned to the robot's flange, resulting in bad pose estimations. A calibration of the camera position with respect to the gripper improved our results considerably, but translational ground truth errors of about 1mm remained. In addition, since aligning the parts to the corner of the table was made by human assessment, we estimated that a ground truth noise of ~0.5 degree exists for the rotational angle.

*B. Experiment*

Nine fixed camera positions in the height of 0.53m were chosen to take images of the parts, which were randomly placed on the table in reachable distance to the robot. For each image, stage 1 of the algorithm was initiated. Since there was an overlap between the nine images, some parts were detected in several images, which required some filtering. The first filter removed small parts (shafts) near the edges of images, since parts detected near the image center enable more accurate image-to-world de-projection, leading to more accurate world pose estimates. The second filter found parts from different classes with similar predicted positions and chose the part with the highest stage 1 classification score. In this process, we removed all misclassified parts as they always had lower classification scores than the true parts. The third filter removed duplicates of the same part by choosing the detection with the highest stage 1 classification score. This filter, which implicitly relies on the assumption of a single part from each class, was only applied once when the *base plate* was detected in two different locations. Next, stage 2 of the algorithm was initiated, resulting in high-accuracy pose estimations used for the assembly.

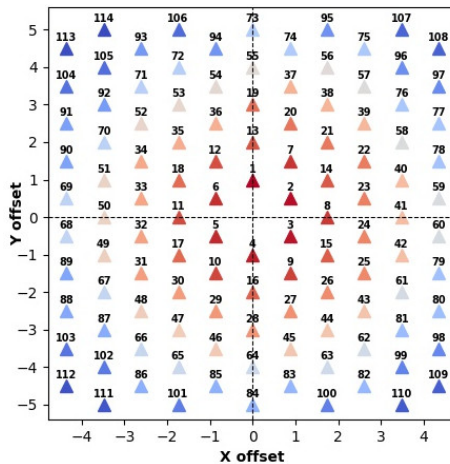

**Figure 5**: Search locations represented as *X* and *Y* offsets in units of mm from the calculated placement position, where numbers and colors indicate the order of search. The size of the search area is 10mm x 10mm and the side length of the triangle d = 1mm, which is the difference between the diameter of the central hole of the compound gear and the diameter of the shaft.

## V. RESULTS

### A. Robotic assembly in a real environment

We performed a total of 58 real experiments. For each experiment, all six parts had to be detected and subjected to pose estimation. Five out of six parts (all but the *base plate*) were then assembled. For cases where the robot failed to assemble a part, the parts were manually assembled, and the robot was then allowed to continue the assembly of the remaining parts. This was done in order to collect assembly success rates for all the parts from a sample of the same size, while success rates in the entire sequence were also measured. As shown in Table 1, our algorithm achieved 100% detection rate. Pose estimation resulted in an average translational accuracy of 2.16mm and a rotational accuracy of 0.64 degrees. Out of 290 attempts of grasp-and-assemble, 264 were successful (91%). In 39 experiments, all parts were successfully assembled (67.2%). The force feedback routine was invoked in only 20% of the cases. When activated, the search algorithm was used with an average of 9.6 location search attempts.

Among the 26 failures, 22 assembly attempts (7.6%) failed due to imperfect grasps using the SAKE EZGripper, which generated too few contact points resulting in too weak grip forces. In 6 of the 22 failed attempts, the gripper succeeded to grasp the part but let it slip while lifting. In the remaining 16 attempts, the grasped part made contact with the *base plate* and the robot sensed a negative feedback force, which triggered the search algorithm for the correct placement position, during which the part slipped. The last 4 failed attempts (1.4%) occurred when the *base plate* was randomly placed in a certain position for which the motion planner could not find a feasible trajectory to place the part. In summary, the angular SAKE EZGripper imposed a severe constraint on the performance. We believe that by using a parallel gripper the grasping and assembly success rates could have been significantly improved, but this needs to be analyzed in future studies.

For comparison of our results with previous works we refer to two related studies. First, we were able to improve pose estimation by an order of magnitude when compared to [5], which also used training in simulation, but used RGB images of more basic parts. Second, we demonstrate assembly capabilities on more parts than considered in [11], which studied a subtask of the *Siemens Innovation Challenge* by inserting a shaft into a gear under fixed initial conditions.

### B. Robotic assembly in simulation

In Table 2 we report our pose estimation results on simulated images. The test set for stage 1 consisted of 30,000 images from a distance of 0.53m (~70,000 parts), whereas 100,000 test images from a distance of 0.31m (100,000 parts) were used for stage 2. As can be seen, our networks achieve near-perfect pose estimation results for simulated images, which are 54 × better than the real results for translations and 7 × better for rotational accuracy. This discrepancy between simulated and real images reflects the remaining simulation-reality gap, which we were not able to close.

TABLE 1. POSE ESTIMATION ON REAL IMAGES AND REAL TASK SUCCESS RATE

| Part Identity | Stage 1 | | | Stage 2 | | | Task Success Rate | |
|---|---|---|---|---|---|---|---|---|
| | *Detection rate* | *Translation accuracy (mm)* | *Rotation accuracy (deg)* | *Translation accuracy (mm)* | *Isometry classification* | *Rotation accuracy (deg)* | *Grasping Success Rate* | *Assembly Success Rate* |
| Base_Plate | 100.00% | 3.73 (2.72) | 1.27 (1.24) | 3.03 (2.48) | - | 0.52 (0.46) | - | - |
| Shaft_1 | 100.00% | 3.72 (2.72) | 108.34 (87.73) | 2.20 (1.20) | 100.00% | 0.65 (0.39) | 100% | 96.55% |
| Shaft_2 | 100.00% | 3.88 (2.37) | 95.12 (89.72) | 1.99 (0.82) | 100.00% | 0.69 (0.42) | 98.28% | 86.21% |
| Compound_Gear | 100.00% | 2.99 (1.64) | 0.61 (0.53) | 2.26 (1.05) | - | 0.71 (0.49) | 98.28% | 98.28% |
| Gear_1 | 100.00% | 2.86 (1.71) | 76.17 (56.65) | 1.77 (0.68) | 100.00% | 0.83 (0.76) | 93.10% | 86.21% |
| Gear_2 | 100.00% | 2.50 (1.86) | 0.70 (0.44) | 1.73 (0.90) | - | 0.44 (0.35) | 100% | 87.93% |
| Average | 100.00% | 3.28 (2.17) | 47.04 (39.39) | 2.16 (1.19) | - | 0.64 (0.48) | 97.93% | 91.04% |

TABLE 2. POSE ESTIMATION ON SIMULATED IMAGES

| Part Identity | Stage 1 | | | Stage 2 | | |
|---|---|---|---|---|---|---|
| | *Detection Rate* | *Translation Accuracy (mm)* | *Rotation Accuracy (deg)* | *Translation Accuracy (mm)* | *Isometry Classification* | *Rotation Accuracy (deg)* |
| Base_Plate | 100.00% | 0.12 (0.07) | 0.28 (0.37) | 0.05 (0.03) | - | 0.03 (0.03) |
| Shaft_1 | 100.00% | 0.18 (0.13) | - | 0.04 (0.03) | 100.00% | 0.04 (0.03) |
| Shaft_2 | 100.00% | 0.17 (0.11) | - | 0.04 (0.02) | 100.00% | 0.04 (0.03) |
| Compound_Gear | 100.00% | 0.11 (0.07) | 0.09 (0.07) | 0.04 (0.03) | - | 0.05 (0.05) |
| Gear_1 | 100.00% | 0.17 (0.11) | - | 0.05 (0.03) | 100.00% | 0.17 (0.16) |
| Gear_2 | 100.00% | 0.11 (0.07) | 0.17 (0.15) | 0.04 (0.03) | - | 0.06 (0.05) |
| Average | 100.00% | 0.13 (0.10) | - | 0.04 (0.03) | - | 0.09 (0.12) |

**Table 1:**
Detection, pose estimation and assembly rate accuracy for parts of the *Siemens Innovation Challenge* in real.

**Table 2:**
Detection and pose estimation in simulation.

**(Both tables):**
For pose estimation, standard deviations are reported in parentheses. Isometry-breaking classification is only relevant for *gear 1*, *shaft 1* and *shaft 2*, which have disturbing near isometries.

## VI. Conclusion

Our work includes several contributions. First, we show that for an assembly task with given 3D CAD files, an accurate pose estimation algorithm can be trained to achieve high assembly success rates. To the best of our knowledge this is the first time that the *Siemens Innovation Challenge* has been addressed with all the parts involved and with high successful assembly rates. Second, we show that a two-stage method - including a stage bringing the sensor close to the part in a canonical pose - can significantly boost pose estimation accuracy. Third, the simulation-to-reality gap can be significantly bridged by using depth images and the assembly task can be learned from simulated images only. Beyond this work, we believe that the presented two-stage pose estimation algorithm is quite generic and can be applied with some adjustments to general assembly tasks. In adapting to a new task, datasets generation and network training in our method can be performed significantly faster than currently available in industrial procedures. It hence has the potential for enabling more flexible manufacturing.